\documentclass[10pt,twocolumn,letterpaper]{article}

\usepackage{wacv}
\usepackage{times}
\usepackage{epsfig}
\usepackage{graphicx}
\usepackage{amsmath}
\usepackage{amssymb}
\usepackage{enumitem}
\usepackage{caption}
\usepackage{xcolor}
\usepackage{colortbl}
\usepackage{caption}
\usepackage{subfigure}
\usepackage{mathrsfs}

\pdfoutput=1

\newcommand{\formattedparagraph}[1]{\noindent \textbf{#1}}

\def\wacvPaperID{135} 

\wacvfinalcopy 

\ifwacvfinal
\fi

\ifwacvfinal
\usepackage[breaklinks=true,bookmarks=false]{hyperref}
\else
\usepackage[pagebackref=true,breaklinks=true,colorlinks,bookmarks=false]{hyperref}
\fi

\begin{document}
\title{Neural Architecture Search for Efficient Uncalibrated Deep Photometric Stereo}

\author{Francesco Sarno$^1$, Suryansh Kumar$^1$,  Berk Kaya$^1$, Zhiwu Huang$^{1}$, Vittorio Ferrari$^{2}$, Luc Van Gool$^{1, 3}$\\
Computer Vision Lab, ETH Z\"urich${^1}$, Google Research$^2$, KU Leuven$^3$
}

\maketitle

\begin{abstract}
We present an automated machine learning approach for uncalibrated photometric stereo (PS). Our work aims at discovering lightweight and computationally efficient PS neural networks with excellent surface normal accuracy. Unlike previous uncalibrated deep PS networks, which are handcrafted and carefully tuned, we leverage differentiable neural architecture search (NAS) strategy to find uncalibrated PS architecture automatically. We begin by defining a discrete search space for a light calibration network and a normal estimation network, respectively. We then perform a continuous relaxation of this search space and present a gradient-based optimization strategy to find an efficient light calibration and normal estimation network. Directly applying the NAS methodology to uncalibrated PS is not straightforward as certain task-specific constraints must be satisfied, which we impose explicitly. Moreover, we search for and train the two networks separately to account for the Generalized Bas-Relief (GBR) ambiguity. Extensive experiments on the DiLiGenT dataset show that the automatically searched neural architectures performance compares favorably with the state-of-the-art uncalibrated PS methods while having a lower memory footprint.

\end{abstract}



\section{Introduction}
\label{sec:intro}
Photometric stereo (PS) aims at recovering an object's surface normals from its light varying images captured from a fixed viewpoint. Although range scanning methods \cite{nayar2012diffuse, kutulakos2008theory, newcombe2011kinectfusion, menini2021real}, multi-view methods \cite{furukawa2009accurate, kumar2017monocular, kumar2019superpixel, kumar2019jumping, kumar2020non, kumar2018scalable} and single image dense depth estimation methods \cite{saxena2008make3d, fu2021single, kumar2019dense} can recover the object's surface normals, photometric stereo is excellent at capturing high-frequency surface details such as scratches, cracks, and dents from images. Therefore, it is a favored choice for fine-detailed surface recovery in many scientific and engineering areas such as forensics \cite{sakarya2008three} and molding \cite{xie2015photometric}. 


Seminal work on PS assumes a Lambertian object under calibrated setting \ie, the directions of the light sources are known \cite{woodham1980photometric}. Firstly, the Lambertian object assumption does not hold for surfaces with general reflectance property. As a result, several robust methods \cite{wu2010robust, oh2013partial, ikehata2014photometric}, and realistic Bidirectional Reflectance Distribution Function (BRDF) based methods \cite{georghiades2003incorporating, chung2008efficient, goldman2009shape, ikehata2014photometricisotropic, shi2013bi} were proposed. Robust methods treat non-Lambertian effects as outliers, and popular realistic BRDF models confine to isotropic BRDF modeling of non-Lambertian surfaces \cite{Ikehata_2014_CVPR, shi2013bi}. Hence, these methods can only model the reflectance property of a restricted class of materials. In general, modeling surfaces with unknown reflectance properties is challenging.


In recent years, deep neural networks have significantly improved the performance of many computer vision tasks, including photometric stereo. Their powerful ability to learn from data has helped in modeling surfaces with unknown reflectance properties, which was a challenge for traditional PS methods. Further, neural networks can implicitly learn the image formation process and global illumination effects from data, which classical algorithms cannot pursue. As a result, several deep learning architectures were proposed for PS \cite{ikehata2018cnn, taniai2018neural, chen2018psfcn, chen2020deep, logothetis2020px, logothetis2020cnn, santo2020deep}. Hence, by leveraging a deep neural network, we can overcome the shortcoming of PS due to the Lambertian object assumption. However, these methods still rely on the other assumption of calibrated setting \ie, the light source directions are given at test time, limiting their practical application. Accordingly, uncalibrated deep PS methods that can provide results comparable to calibrated PS networks are becoming more and more popular \cite{chen2019self, kaya2020uncalibrated, chen2020learned}.

The impressive results demonstrated by deep uncalibrated PS methods have a few critical issues: the network architecture is manually designed, and therefore, such networks are typically not optimally efficient and have a large memory footprint \cite{kaya2020uncalibrated, chen2019self, chen2018psfcn, chen2020learned}. Moreover, the authors of such networks conduct many experiments to explore the effect of empirically selected operations and tune hyperparameters. But, we know from the popular research in machine learning that not only the type of operation but sometimes their placement (ordering) matters for performance \cite{he2016identity, yao2020gps}. And therefore, a separate line of research known as Neural Architecture Search (NAS) has gained tremendous interest to tackle such challenges in architecture design. NAS methods automate the design process, greatly reducing human effort in searching for an efficient network design \cite{AUTOML}. NAS algorithms have shown great success in many high-level computer vision tasks such as object detection \cite{tan2020efficientdet, xu2019auto}, image classification \cite{wu2019fbnet}, image super-resolution \cite{wu2021trilevel}, action recognition \cite{sukthanker2020neural}, and semantic segmentation \cite{liu2019auto}. Yet, its potential for low-level 3D computer vision problem such as uncalibrated PS remains unexplored. 

Among architecture search methods, evolutionary algorithms \cite{real2019regularized, real2017large} and reinforcement learning-based methods \cite{zoph2016neural, zoph2018learning} are computationally expensive and need thousands of GPU hours to find architecture. Hence they are not suitable for our problem. Instead, we adhere to the cell-based differentiable NAS formulation. It has proven itself to be computationally efficient and demonstrated encouraging results for many high-level vision problems \cite{liu2018darts, liu2019autodeeplab}. However, in those applications, differentiable NAS is used without any task-specific treatment. Unfortunately, this will not work for the uncalibrated PS problem.  There exists GBR ambiguity \cite{belhumeur1999bas} due to the lack of light source information.
Moreover, certain task-specific constraints must be satisfied (\eg, unit normal, unit light source direction), and the method must operate on unordered image sets. Unlike typical NAS-based methods, we incorporate human knowledge in our search strategy to address those challenges. To resolve GBR ambiguity, we first search for an efficient light calibration network, followed by a normal estimation network's search \cite{chen2019self}.  To handle PS-related constraints, we fix some network layers and define our discrete search space for both networks accordingly. We model our PS architecture search space via a continuous relaxation of the discrete search space, which can be optimized efficiently using a gradient-based algorithm.

We evaluated our method's performance on the DiLiGenT benchmark PS dataset \cite{shi2016benchmark}. The experiments reveal that our approach discovers lightweight architectures, which provides results comparable to the state-of-the-art manually designed deep uncalibrated networks \cite{chen2018psfcn, chen2019self, kaya2020uncalibrated}. This paper makes the following contributions:
%
\begin{itemize}[leftmargin=*, noitemsep]
\item We propose the first differentiable NAS-based framework to solve uncalibrated photometric stereo problem.
\item Our architecture search methodology considers the task-specific constraints of photometric stereo during search, train, and test time to discover meaningful architecture.
\item We show that automatically designed architecture outperforms the existing traditional uncalibrated PS performance and compares favorably against hand-crafted deep PS network with significantly less parameters.
\end{itemize}






\section{Proposed Method}
This section describes our task-specific neural architecture search (NAS) approach. We utilize the seminal classical photometric stereo formulation \cite{woodham1980photometric, belhumeur1999bas} and previous handcrafted deep neural network design \cite{chen2019self} as the basis of our NAS framework. Utilizing previous methods knowledge in the architecture design process not only helps in reducing the architecture's search time but also provides an optimal architecture with better performance accuracy \cite{chen2019self, kaya2020uncalibrated}. Before we describe the NAS modeling of our problem, we define the classical photometric stereo setup.

Consider an orthographic camera observing a rigid object from a given viewpoint $\mathbf{v} = (0, 0, 1)^{T}$. For PS setup, the images are captured by firing one unique directional light source per image. Let $I \in \mathbb{R}^{m \times n}$ be the measurement matrix comprising of $n$ images with $m$ pixels stacked as column vectors. Let $L \in \mathbb{R}^{3 \times n}$ and $N \in \mathbb{R}^{3 \times m}$ denote all the light sources and surface normals respectively.  Then, the image formation model under Lambertian surface assumption is formulated as follows:

\begin{equation}\label{eq:classicalPS}
    I = \rho \cdot N^{T}L~+~E .
\end{equation}
\noindent
Here, $\rho \in \mathbb{R}$ is the diffuse albedo and $E$ accounts for error due to shadows, specularities, or noise. When all non-Lambertian effects are ignored, solving Eq.\eqref{eq:classicalPS} can recover the actual surface up to a GBR transformation, such that $I = (G^{-T}\bar{N})^{T}(GL)$. Here, $\bar{N} \in \mathbb{R}^{3 \times m}$ denotes the albedo scaled normals and $G \in \mathbb{R}^{3 \times 3}$ is the transformation matrix with 3 unknown parameters \cite{belhumeur1999bas, chandraker2005reflections}. It indicates that there are many solutions leading up to the same image. Nevertheless, it is well known that specularities \cite{georghiades2003incorporating, drbohlav2005can}, interreflections \cite{chandraker2005reflections}, albedo distributions \cite{alldrin2007resolving, shi2010self} and BRDF properties \cite{tan2007isotropy, wu2013calibrating, lu2017symps} provide useful cues for disambiguation. However, such cues are not well exploited in a single-stage network designed for regressing per-pixel normals, and therefore, we adhere to use two different neural networks following Chen \etal \cite{chen2019self}. 
We first learn the light sources from images by training a light calibration network in a supervised way. Then, we use its results at inference time for the normal estimation network to predict the surface normals. Unlike other uncalibrated deep PS methods, our approach allows automatic search for the optimal architecture both for the light calibration and normal estimation networks.


\subsection{Architecture Search for Uncalibrated PS}
Leveraging the recent one-shot cell-based NAS method \ie, DARTS \cite{liu2018darts}, we first define different discrete search spaces for light calibration and normal estimation networks. Next, we perform a continuous relaxation of these search spaces, leading to differentiable bi-level objectives for optimization. We perform an end-to-end architecture search for light calibration and normal estimation networks separately to obtain optimal architectures. Contrary to high-level vision problems such as object detection, image classification, and others \cite{liu2018darts, chu2020fair, zoph2018learning}, directly applying the one-shot NAS to existing uncalibrated PS networks \cite{chen2019self, kaya2020uncalibrated} may not necessarily lead to a good solution. Unfortunately, for our task, a single end-to-end NAS seems challenging. It may lead to unstable behavior due to GBR ambiguity \cite{belhumeur1999bas}. And therefore, we search for an optimal light calibration first and then search for a normal estimation network by keeping some of the necessary operations or layers fixed ---such a strategy is used in other NAS based applications \cite{fu2020autogan}. The searched architectures are then trained independently for inference.

\formattedparagraph{$\bullet$ Background on Differentiable NAS.} 
In recent years, Neural Architecture Search (NAS) has attracted a lot of attention from the computer vision research community. The goal of NAS is to automate the process of deep neural network design. Among several promising approaches proposed in the past \cite{real2019regularized, zoph2018learning, real2019regularized, liu2018darts, liu2017progressive, kandasamy2019neural, pham2018efficient}, the DARTS \cite{liu2018darts} has shown promising outcomes due to its computational efficiency and differentiable optimization formulation. So, in this paper, we use it to design an efficient deep neural network to solve uncalibrated PS.

\begin{figure}[t]
\centering
\subfigure[\label{fig:darts_stage_0}]{\includegraphics[width=0.10\textwidth]{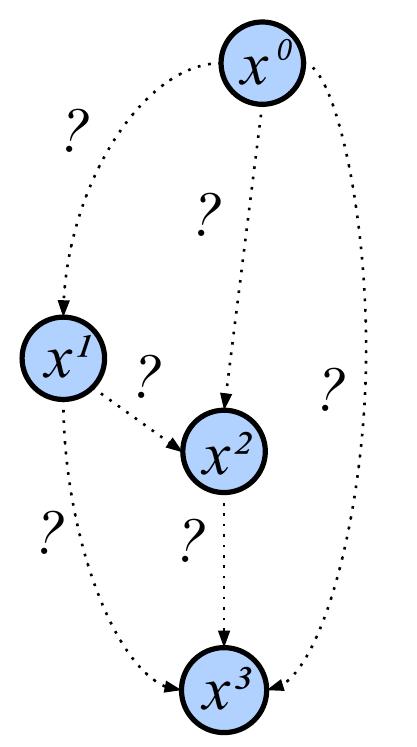}}~
\subfigure[\label{fig:darts_stage_1}]{\includegraphics[ width=0.10\textwidth, ]{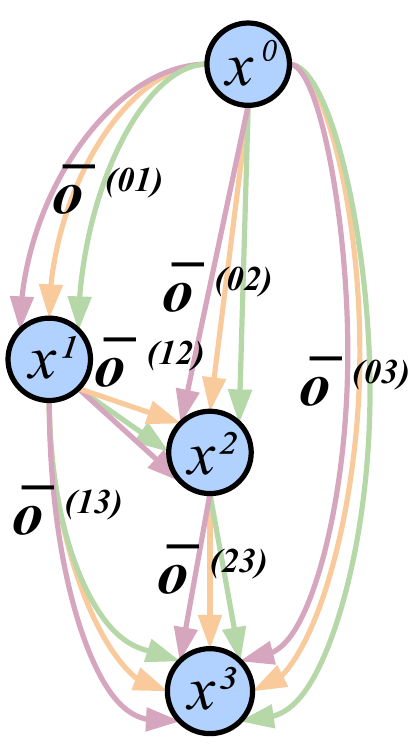}}~~
\subfigure[\label{fig:darts_stage_2}]{\includegraphics[ width=0.10\textwidth, ]{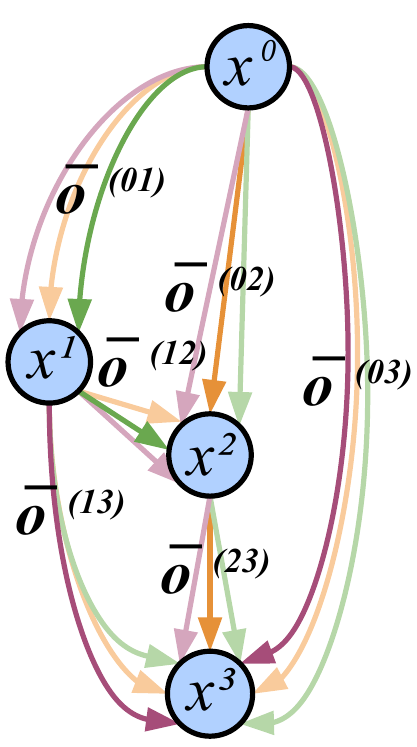}}~~
\subfigure[\label{fig:darts_stage_3}]{\includegraphics[ width=0.10\textwidth, ]{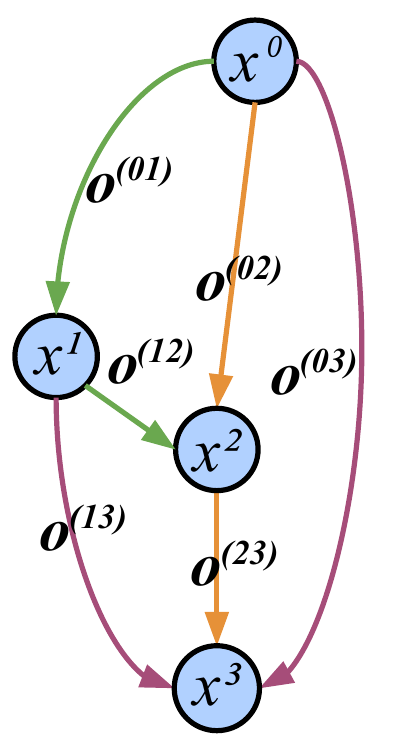}}~~
\caption{\small 
Illustration of a cell. (a) Initially, the optimal operations $\bar{o}^{(i, j)}$  between nodes $x^{(i)}$ and $x^{(j)}$ are unknown. (b) Each node is computed by a mixture of candidate operations.
(c) Architecture encoding is obtained by solving the continuous relaxation of the search space. (d) Optimal cell obtained after selection of most probable candidate operation.
}
\label{fig:darts_cell_evolution}
\end{figure}

DARTS searches for a computational cell from a set of defined search spaces, which is a building block of the architecture. Once the optimal cells are obtained, it is stacked to construct the final architecture for training and inference. To find the optimal cell, we define search space $\mathcal{O}$, that is a set of possible candidate operations. The method first performs continuous relaxation on the search spaces and then searches for an optimal cell. A cell is a directed acyclic graph (DAG) with $N$ nodes and $E$ edges. Each node is a latent feature map representation say $x^{(i)}$ for the $i^{th}$ node and each edge is associated with an operation say $o^{(i, j)}$ between node $i$ and node $j$ (see Fig.\ref{fig:darts_stage_0}). In a cell, each intermediate node is computed from its preceding nodes as follows: 
%
\begin{equation}
    x^{(j)} = \sum_{i < j} o^{(i, j)} \big(x^{(i)}\big)
\end{equation}
\noindent
Let $o^{(i, j)}$ be some operation among $K$ candidate operations $\mathcal{O} = \{o^{(i,j)}_1, o^{(i,j)}_2, ..., o^{(i,j)}_K\}$. The categorical choice of a specific operation is replaced by the continuous relaxation of the search space by taking softmax over all the defined candidate operations as follows:
%
\begin{equation}
    \Tilde{o}^{(i, j)}(x) = \sum_{o \in \mathcal{O}} \frac{\exp(\alpha_o^{(i, j)})}{\sum_{o' \in \mathcal{O}} \exp(\alpha_{o'}^{(i, j)})} o(x)
    \label{eq:cont_relax}
\end{equation}
%
%
Here, $\alpha^{(i,j)}$ is a vector of dimension $|\mathcal{O}|$  which denotes the operation mixing weights on edge ${(i,j)}$ (see Fig.\ref{fig:darts_stage_1}). As a result, the search task for DARTS reduces to a learning set of continuous variable $\alpha^{(i, j)} ~\forall ~(i, j)$. The optimal architecture will be determined replacing each mixed operation $\bar{o}^{(i,j)}$ on edge ${(i,j)}$ with: ${o}^{(i,j)} = {\arg\max}_{o \in \mathcal{O}} \; {\alpha_{o}^{(i,j)}}$ corresponding to the operation which is the ``most probable'' among the ones listed in $\mathcal{O}$ (see Fig.\ref{fig:darts_stage_2}-Fig.\ref{fig:darts_stage_3}). The introduced relaxation allows joint learning of architecture $\alpha$ and its weight $\omega$ within the mixture of operations. So, the goal of architecture search now becomes to search for an optimal architecture $\alpha$ using the validation loss with the weights $\omega$ that minimizes the training loss for a given $\alpha$. This leads to following bi-level optimization problem.
\begin{equation}
\begin{aligned}
     & \displaystyle  \underset{\alpha}{\textrm{minimize}}  ~~\mathscr{L}_{val}(\omega^{*}(\alpha), \alpha);\\
    & \displaystyle  \textrm{subject to:} ~\omega^{*}(\alpha) = \underset{\omega}{\arg\min} ~~\mathscr{L}_{train}(\omega, \alpha)
\label{eq:bi_level_opt}
\end{aligned}
\end{equation}
where, $\mathscr{L}_{val}$ and $\mathscr{L}_{train}$ are the validation and training losses respectively. This optimization problem is solved iteratively until convergence is reached. The architecture $\alpha$ is updated by substituting the lower-level optimization gradient approximation. Concretely, update $\alpha$ by descending $\nabla_{\alpha}\mathscr{L}_{val}(\omega-\xi\nabla_{\omega}\mathscr{L}_{train}(\omega, \alpha), \alpha)$. Subsequently update $\omega$ by descending $\nabla_{\omega} \mathscr{L}_{train}(\omega,\alpha)$, where:
\begin{equation}
\begin{aligned}
    \nabla_{\alpha} \mathscr{L}_{val}(\omega - \xi\nabla_{\omega}\mathscr{L}_{train}(\omega, \alpha), \alpha) \approx \nabla_{\alpha} \mathscr{L}_{val}(\omega^{*}(\alpha),\alpha)
\end{aligned}\label{eq:second_order_approx}
\end{equation}
$\xi > 0$ is the learning rate of the inner optimization. The idea is that, $\omega^{*}(\alpha)$ is approximated with a single learning step which allows the searching process to avoid solving the inner optimization in Eq.\eqref{eq:bi_level_opt} exactly. We refer this formulation as second-order approximation \cite{liu2018darts}. To speed up the searching process, common practice is to apply first-order approximation by setting $\xi = 0$. For more details on the bi-level optimization refer Liu \etal work \cite{liu2018darts}.

\vspace{-0.1cm}

\subsubsection{Our Cell Description}\label{sss:cell_description}
For our problem, we search for both light calibration and normal estimation networks. Our cells consist of two input nodes, four intermediate nodes, and one output node for both of the networks. Each cell at layer $k$ uses the output of two preceding cells ($\mathcal{C}_{k-1}$ and $\mathcal{C}_{k-2}$) at input nodes and outputs $\mathcal{C}_{k}$ by channel-wise concatenation of the features at the intermediate nodes. To adjust the spatial dimensions, we define two cells \ie, \emph{normal cells} and \emph{reduction cells}. Normal cells preserve the spatial dimensions of the input feature maps by applying convolution operations with stride 1. The reduction cells use operations with stride 2 adjacent to input nodes, reducing the spatial dimension by half. Although the cell definition for both networks is the same, the network-level search spaces are different due to the problem's constraints. Next, we describe our procedure to obtain optimal network architecture for uncalibrated PS.

\subsection{Light Calibration Network}
Light calibration network predicts all the light source's direction and intensity from a set of PS images. Here, we assume the object mask is known. One obvious way to estimate light is to regress a set of images with the source direction vectors and intensities in a continuous space. However, converting this task into a classification problem is more favorable for our purpose. It stems from the fact that learning to classify light source directions to predefined bins of angles is much easier than regressing the unit vector itself. Further, using discretized light directions makes the network robust to small input variations.


We represent the light source direction in the upper-hemisphere by its azimuth $\phi \in [0,\pi]$ and elevation  $\theta \in [-\pi/2,\pi/2]$ angles. We divide the angle spaces into 36 evenly spaced bins $(K_d = 36)$. Our network perform classification on azimuth and elevation separately. For the light intensities, we assign the values in the range of $[0.2, 2]$ divided uniformly into 20 bins $(K_e = 20)$ \cite{chen2019self}.

\begin{figure*}[t]
\centering
\subfigure[\label{fig:LCNet_pipeline} Light Calibration Network ]{\includegraphics[ height=0.18\textwidth]{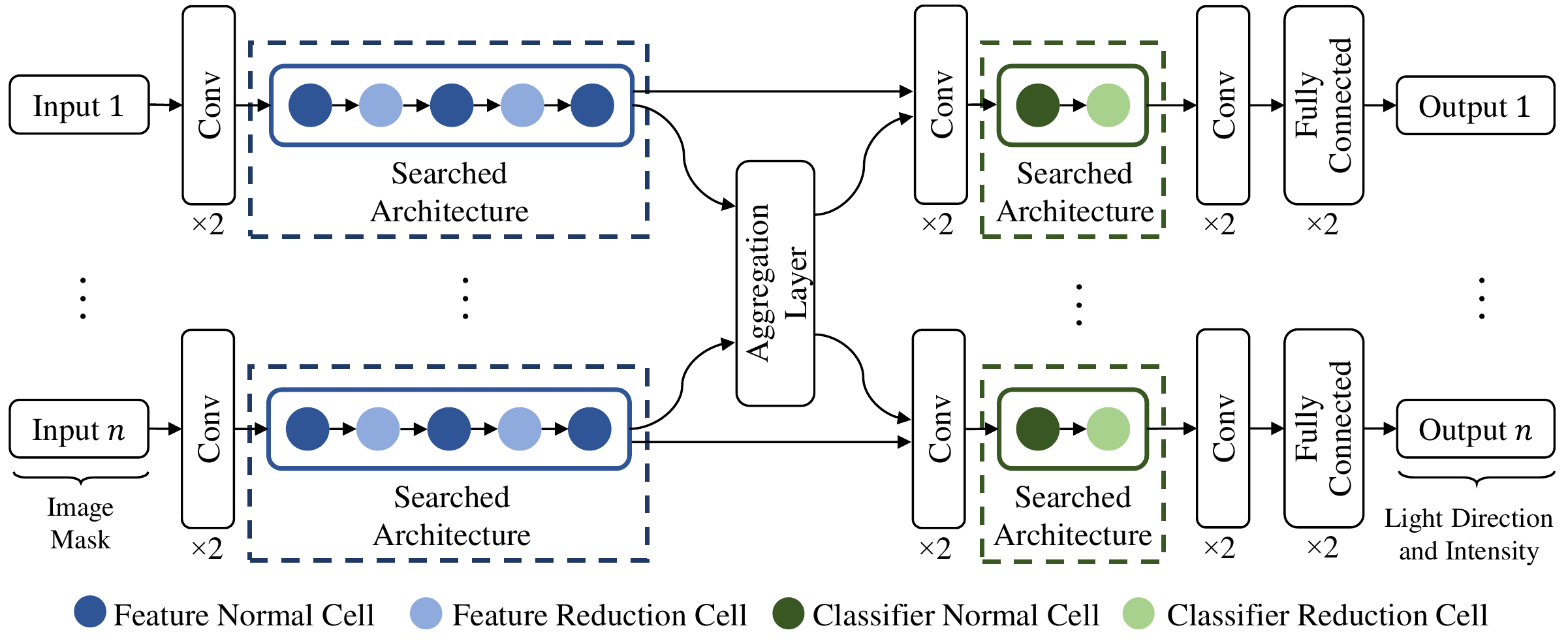}}
~~~~~~~\subfigure[\label{fig:NENet_pipeline} Normal Estimation Network]{\includegraphics[height=0.18\textwidth]{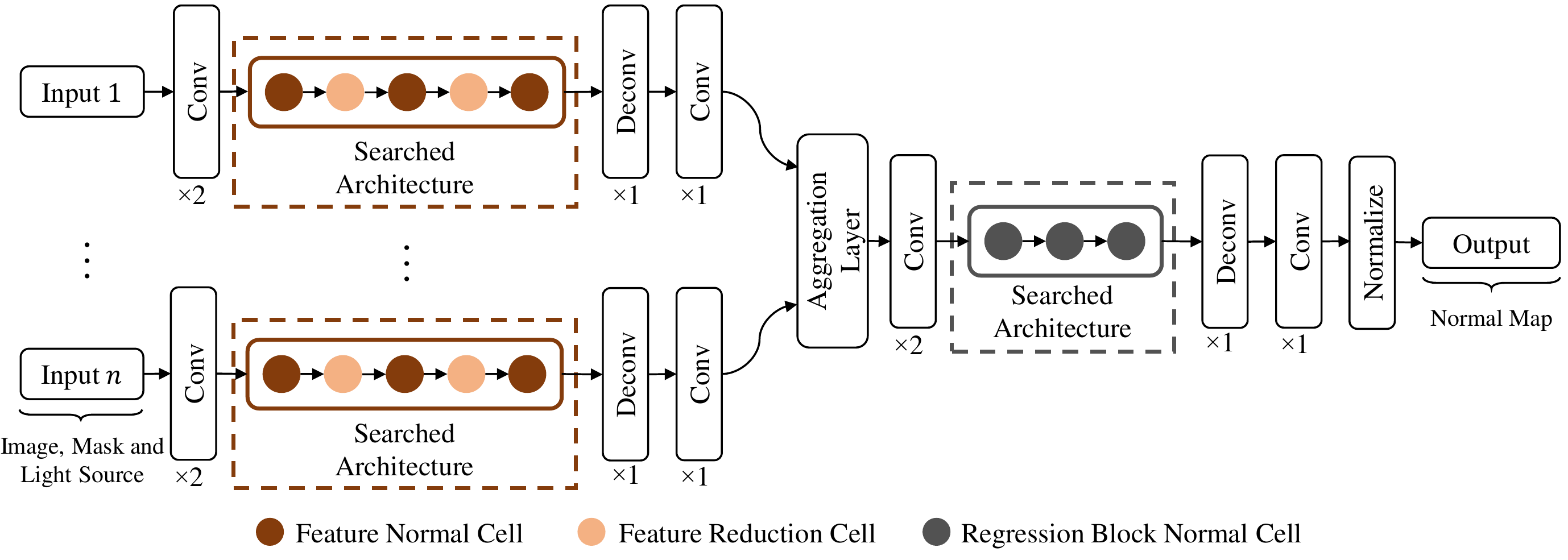}}
\caption{\small Our pipeline consists of two  networks: (a) Light Calibration Network predicts light source directions and intensities from images. Our search is confined to feature extraction module and classification module. 
(b) Normal Estimation Network outputs the surface normal map from images and estimated light sources.
Our search is confined to feature extraction module and regression module.
}
\label{fig:pipeline}
\end{figure*}

\formattedparagraph{NAS for Light Calibration Network.} To perform NAS for light calibration network, we use the backbone shown in Fig.\ref{fig:LCNet_pipeline}. The backbone consists of three main parts $(i)$ local feature extractor $(ii)$ aggregation layer and $(iii)$ classifier. The feature extraction layers provide image-specific information for each input image. The weights of these feature extraction layers are shared among all input images. The image-specific features are then aggregated to a global feature representation with the max-pooling operation. Later, global feature representation is combined with the image-specific information and fed to the subsequent layers for classification. The fully connected layers provide softmax probabilities for azimuth, elevation, and intensity values. 

We use the NAS algorithm to perform search only over the feature extraction layer and classifier layers for architecture search (shown with dashed box Fig.\ref{fig:LCNet_pipeline}), while keeping other layers fixed. For NAS to provide optimal architecture over the searchable blocks in the light-calibration network backbone, we define our search space as follows:

\noindent
\textit{1. Search Space.} Our candidate operations set in search space for light calibration network is composed of $\mathcal{O}^{light} = \{
``1\times1 ~\mathrm{separable~conv.}" , 
``3\times3 ~\mathrm{separable~conv.}" , 
``5\times5 ~\mathrm{separable~conv.}" , 
``\mathrm{skip~connection}",
``\mathrm{zero}"
\}$. The ``$\mathrm{zero}$" operation indicates the lack of connection between two nodes. Each convolutional layer defined in the set first applies ReLU \cite{xu2015empirical} and then convolution with given kernel size followed by batch-normalization \cite{ioffe2015batch}.
As before, our cells consist of two input nodes, four intermediate nodes, and one output node \S \ref{sss:cell_description}. Just for the initial cell, we use stem layers as its input for better search. These layers apply fixed convolutions to enrich the initial cell input features.

\noindent
\textit{2. Continuous relaxation and Optimization.} We perform the continuous relaxation of our defined search space using Eq.\eqref{eq:cont_relax} for differentiable optimization. During searching phase, we perform alternating optimization over weights $\omega$ and architecture encoding values $\alpha$ as follows:

\begin{itemize}[leftmargin=*,topsep=0pt, noitemsep]
    \item Update network weights $\omega$ by $\nabla_{\omega} \mathscr{L}_{train}(\omega, \alpha)$.
    \item Update architecture mixing weights $\alpha$ by $\nabla_{\alpha} \mathscr{L}_{val}(\omega - \xi\nabla_{\omega}\mathscr{L}_{train}(\omega, \alpha), \alpha)$. (see Eq.\eqref{eq:second_order_approx} )
\end{itemize}
$\mathscr{L}_{train}$ and $\mathscr{L}_{val}$ denote the loss computed over training and validation datasets, respectively. We use multi-class cross-entropy loss on azimuth, elevation, and intensity classes to optimize our network \cite{chen2019self}. The total light calibration loss is

\begin{equation}
    \begin{aligned}\label{eq:lcnet_loss}
        \mathscr{L}_{light} = \mathscr{L}_{\phi} +  \mathscr{L}_{\theta} + \mathscr{L}_{e}
    \end{aligned}
\end{equation}
where, $\mathscr{L}_{\phi}$, $\mathscr{L}_{\theta}$, and $\mathscr{L}_{e}$ are the losses for azimuth, elevation, and intensity respectively. We utilize the synthetic Blobby and Sculpture datasets \cite{chen2018psfcn} for this optimization where ground-truth labels for lighting are provided.

Once the searching phase is complete, we convert the continuous architecture encoding values into a discrete architecture. For that, we select the strongest operation on each edge ${(i,j)}$ with: ${o}^{(i,j)} = {\arg\max}_{o \in \mathcal{O}^{light}} \; {\alpha_{o}^{(i,j)}}$. We preserve only the strongest two operations preceding each intermediate node. We train our designed architecture with optimal operations from scratch on the training dataset again to optimize weights before testing \S \ref{sss:search_train_details}.

\subsection{Normal Estimation Network}
We independently search for optimal normal estimation network using the backbone shown in Fig.\ref{fig:NENet_pipeline}. To use the light source information into the network, we first convert $n$ light direction vectors into a tensor $\mathcal{X} \in \mathbb{R}^{n \times 3 \times h \times w }$, where each 3-vector is repeated over spatial dimensions $h$ and $w$. This tensor is then concatenated with the input image to form a tensor $\mathcal{I} \in \mathbb{R}^{n \times 6 \times h \times w }$. Similar to the light calibration network, we use a shared-weight feature extraction block to process each input. After image-specific information is extracted, we combine them in a fixed aggregation layer with the max-pooling operation and obtain a global representation. Keeping the aggregation layer fixed allows the network to operate on an arbitrary number of test images and improves robustness. The global information is finally used to regress the normal map, where a fixed normalization layer is used to satisfy the unit-length constraint.

\formattedparagraph{NAS for Normal Estimation Network.}
Similar to light calibration network, the cells here consist of two input nodes, four intermediate nodes, and one output node. 
To efficiently search for architectures at initial layers, we make use of stem layers prior to each search space \cite{liu2018darts}. These layers apply fixed convolutions to enrich the input features.
 
\noindent
\textit{1. Search Space.} It is a well-known fact that the kernel size has great importance in vision problems. Recent work on photometric stereo has verified that using bigger kernel size helps to explore the spatial information, but stacking too many of them leads to over-smoothing and degrades the performance \cite{yao2020gps}.  
Therefore, we selectively use different kernel sizes in the candidate operations set $\mathcal{O}^{normal} = \{
``1\times1~\mathrm{separable~conv.}" , 
``3\times3~\mathrm{separable~conv.}" , 
``5\times5~\mathrm{separable~conv.}" , 
``\mathrm{skip~connection}",
``\mathrm{zero}"
\}$. Here also, each convolutional layer defined in the set first applies ReLU \cite{xu2015empirical} and then convolution with given kernel size followed by batch-normalization \cite{ioffe2015batch}. 
The selection of candidate operation sets if further investigated in \S 3.3 of the supplementary material.



\noindent
\textit{2. Continuous Relaxation and Optimization.} Similar to light calibration network, we use Eq.\eqref{eq:cont_relax} to make the search space continuous. We then jointly search for the architecture encoding values and the weights using the ground-truth surface normals and light source information during optimization. The optimization is performed using the same bi-level optimization approximation strategy (see Eq.\eqref{eq:bi_level_opt} and Eq.\eqref{eq:second_order_approx}). We normalize the images before feeding them to the network. The normalization ensures the network is robust to different intensity levels. To search normal estimation network, we use the following cosine similarity loss:
 
\begin{equation}
    \mathscr{L}_{normal} = \frac{1}{m} \sum_i^{m}(1-\boldsymbol{\tilde{n}}_i^T \boldsymbol{n}_i) 
    \label{eq:normal_loss}
\end{equation}
where, $\boldsymbol{\tilde{n}}_i$ is the estimated normal by our network and $\boldsymbol{n}_i$ is the ground-truth normal at pixel $i$. Note that $\boldsymbol{\tilde{n}}_i$ is a unit-vector due to the fixed normalization layer.


After the search optimization for normal estimation network is done, we obtain optimal discrete architecture by keeping the operation ${o}^{(i,j)} = {\arg\max}_{o \in \mathcal{O}^{normal}} \; {\alpha_{o}^{(i,j)}}$ on
each edge $(i,j)$. Similar to \cite{liu2018darts}, we only preserve the two preceding operations with highest weight for each node. Finally, we train our normal estimation network from scratch using the searched architecture. Our normal estimation network uses the light directions and intensities estimated by the light calibration network to predict normals at test time.

\begin{figure*}[t]
\centering
\subfigure[\label{fig:LCNet_training_curve} Training Curve of Light Calibration Net]{\includegraphics[width=0.30\textwidth, height=0.22\textwidth]{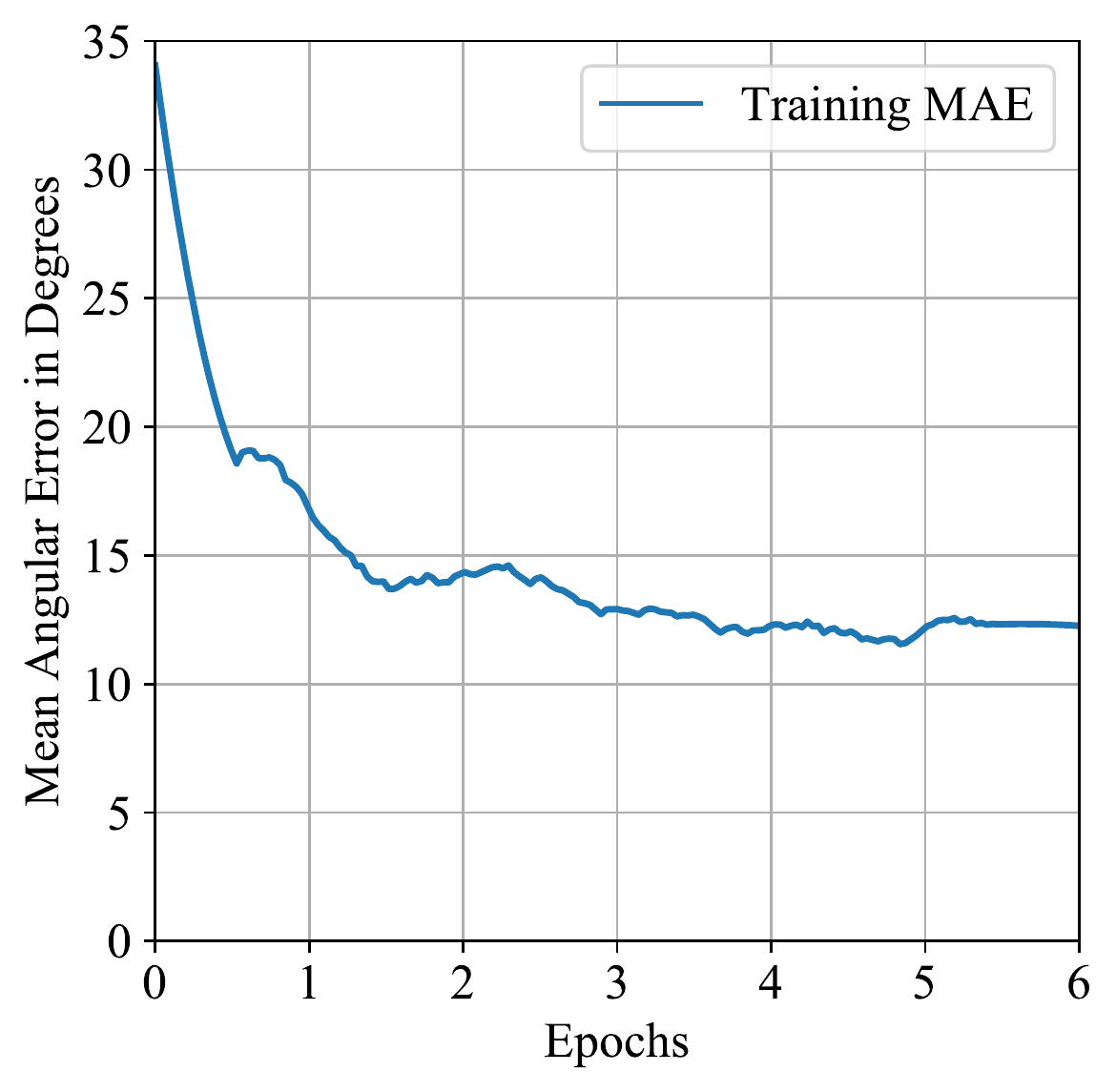}}
~~~~\subfigure[\label{fig:light_maps} Light Directions and Intensities obtained using Light Calibration Network]{
\includegraphics[width=0.50\textwidth, height=0.22\textwidth]{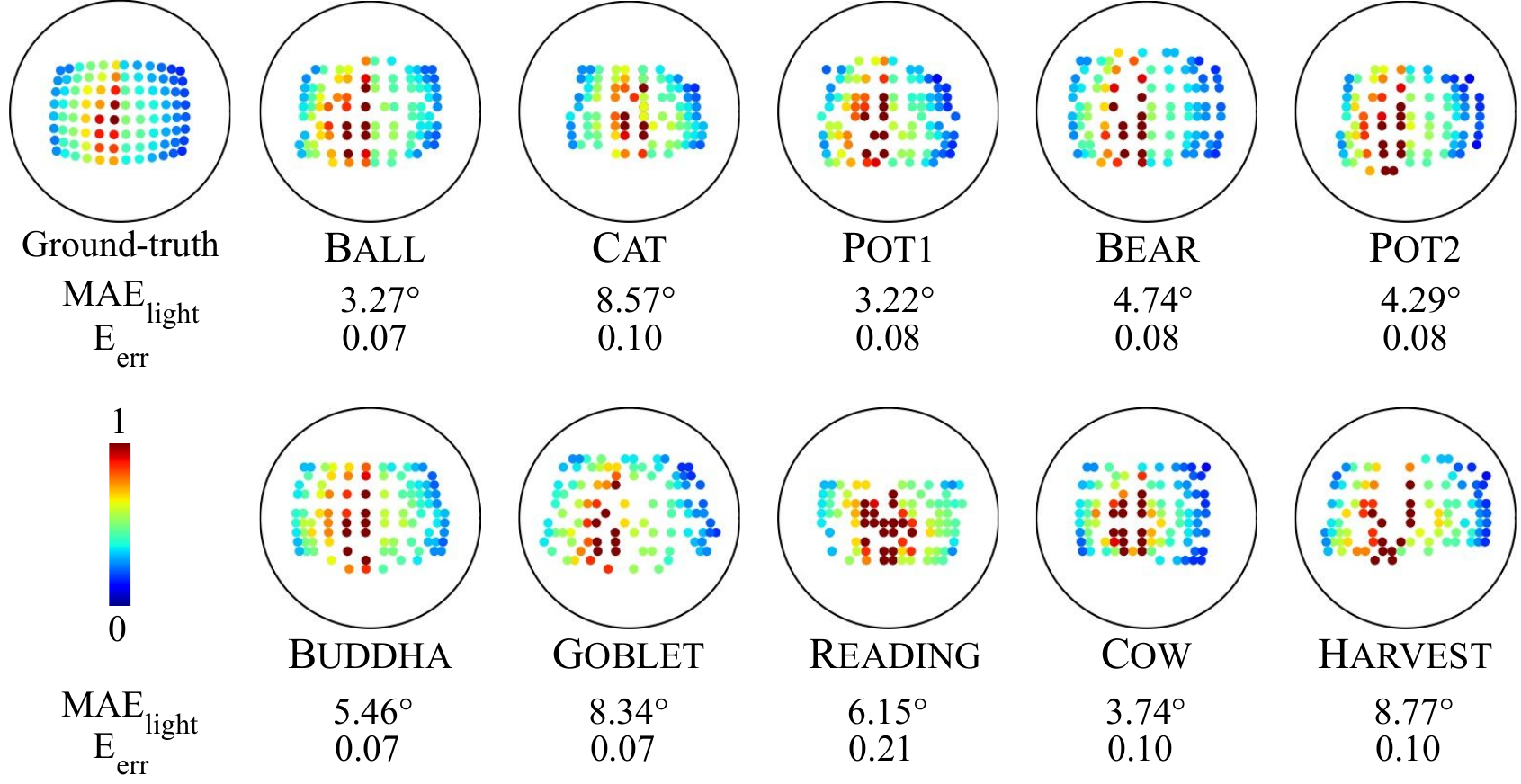}
}
\caption{\small (a) Training curve of the light calibration network.
(b) Light calibration network results on DiLiGenT objects. 
We show the light direction by projecting the vector $[x, y, z]$ to a corresponding point $[x, y]$.
The color of the point shows the light intensity value in $[0, 1]$ range. $\text{MAE}_{light}$ is the mean angular error in the estimation of light source direction and $\text{E}_\text{err}$ stands for the intensity error.
}
\label{fig:light_calibration_results}
\end{figure*}

\section{Experiments and Results}\label{sec:experiments}
This section first describes our procedure in preparing the dataset for the searching, training, and testing phase. Later, we provide the implementation of our method, followed by statistical evaluations and ablation. 

\subsection{Dataset Preparation}
We used three popular photometric stereo datasets for our experiments, statistical analysis, and comparisons, namely, Blobby \cite{10.1109/CVPR.2011.5995510}, Sculpture \cite{wiles2017silnet} , and DiLiGenT \cite{shi2019tpami}.

\smallskip
\formattedparagraph{Search and Train Set Details.}\label{sss:search_train_details}
For architecture search and optimal architecture training, we used 10 objects from the Blobby dataset \cite{10.1109/CVPR.2011.5995510} and 8 from the Sculpture dataset \cite{wiles2017silnet}. We considered the rendered photometric stereo images of these datasets provided by Chen \etal \cite{chen2018psfcn}. It uses 64 random lights to render the objects. In search and train phase, we randomly choose 32 light source images. Following Chen \etal \cite{chen2018psfcn}, we considered $128 \times 128$ sized images for both Blobby and Sculpture dataset.

\smallskip
\formattedparagraph{(a) Preparation of Search Set.} Searching for an optimal architecture using one-shot NAS \cite{liu2018darts} can be computationally expensive. To address that, we use only $10\%$ of the dataset such that it contains subjects from all the categories present in the Blobby and Sculpture dataset. Next, we resized all those $128 \times 128$ resolution images to $64 \times 64$. We refer this dataset as Blobby search set and Sculpture search set. Our search set is further divided into search train set and search validation set. This train set is prepared by taking eight shapes from Blobby search set and six shape from Sculpture search set. The search validation set is composed of two shapes from Blobby and Sculpture search sets, respectively. Hence, approximately $80\%$ of the search set is used as search train set and $20\%$ is used as search validation set. This is done in a way that there is no common subject between train and validation sets. We used a batch size of four at train and validation time during search phase. The search set is same for the light calibration and normal estimation network's search.


\formattedparagraph{(b) Preparation of Train Set.} Once the optimal architectures for light calibration and normal estimation are obtained, we use the train set for training these networks from scratch. Since, we searched architecture using $64 \times 64$ size images, we use 
convolution layer with stride 2 at the train time for the light calibration network's training. Following Chen \etal \cite{chen2018psfcn}, we use $99\%$ of the Blobby and Sculpture dataset for training and $1\%$ for the validation. For light calibration we used batch size of thirty-two at train time and eight for validation. For normal estimation, instead, we considered batch size of four both at training and validation.


\begin{figure*}[t]
\centering
\subfigure[\label{fig:NENet_training_curve} Training Curve of Normal Estimation Net]{\includegraphics[width=0.30\textwidth, height=0.22\textwidth]{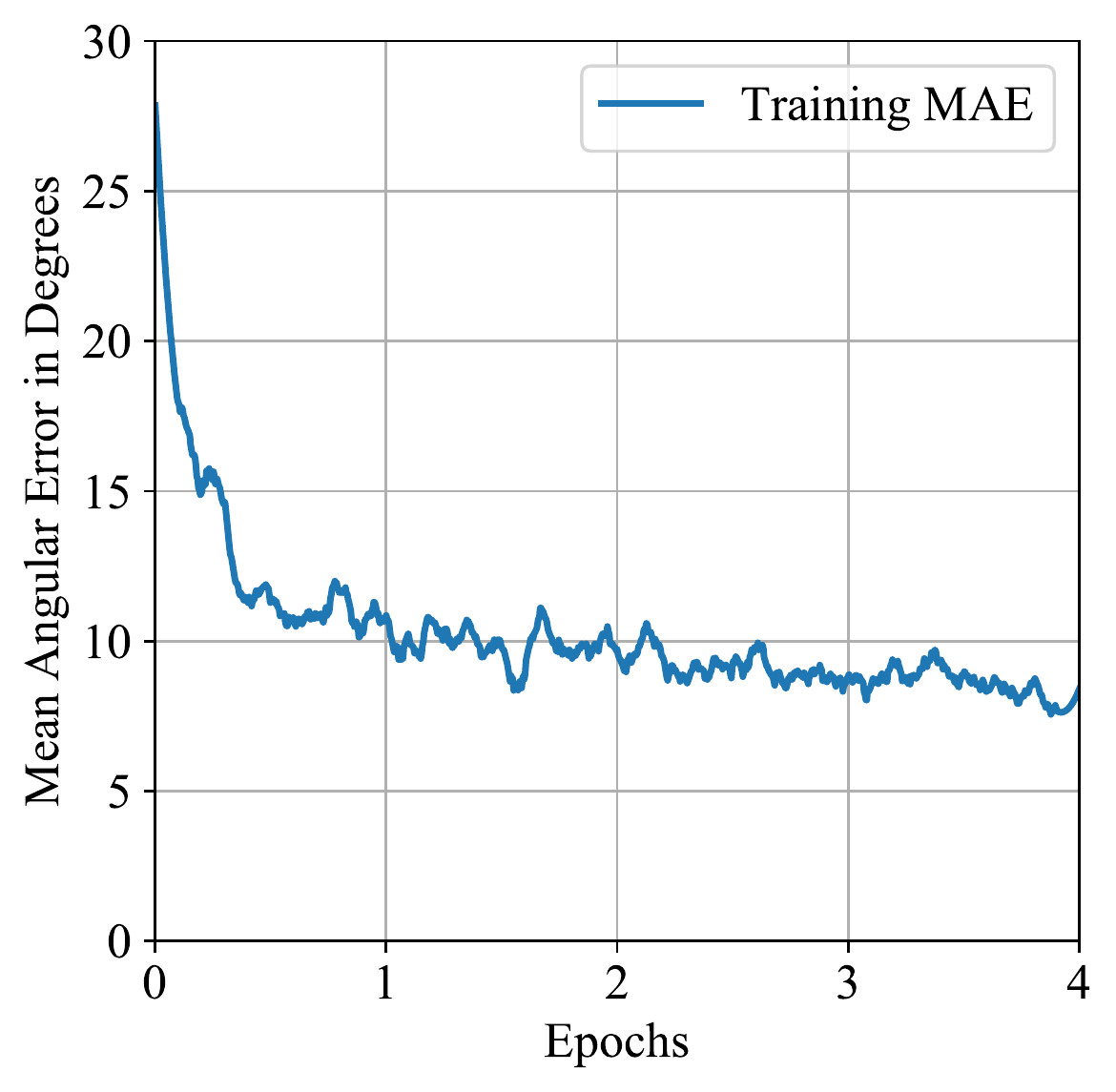}}
~~~\subfigure[\label{fig:diligent_normal_results_ours} Surface Normals obtained using Normal Estimation Network ]{
\includegraphics[width=0.54\textwidth, height=0.22\textwidth]{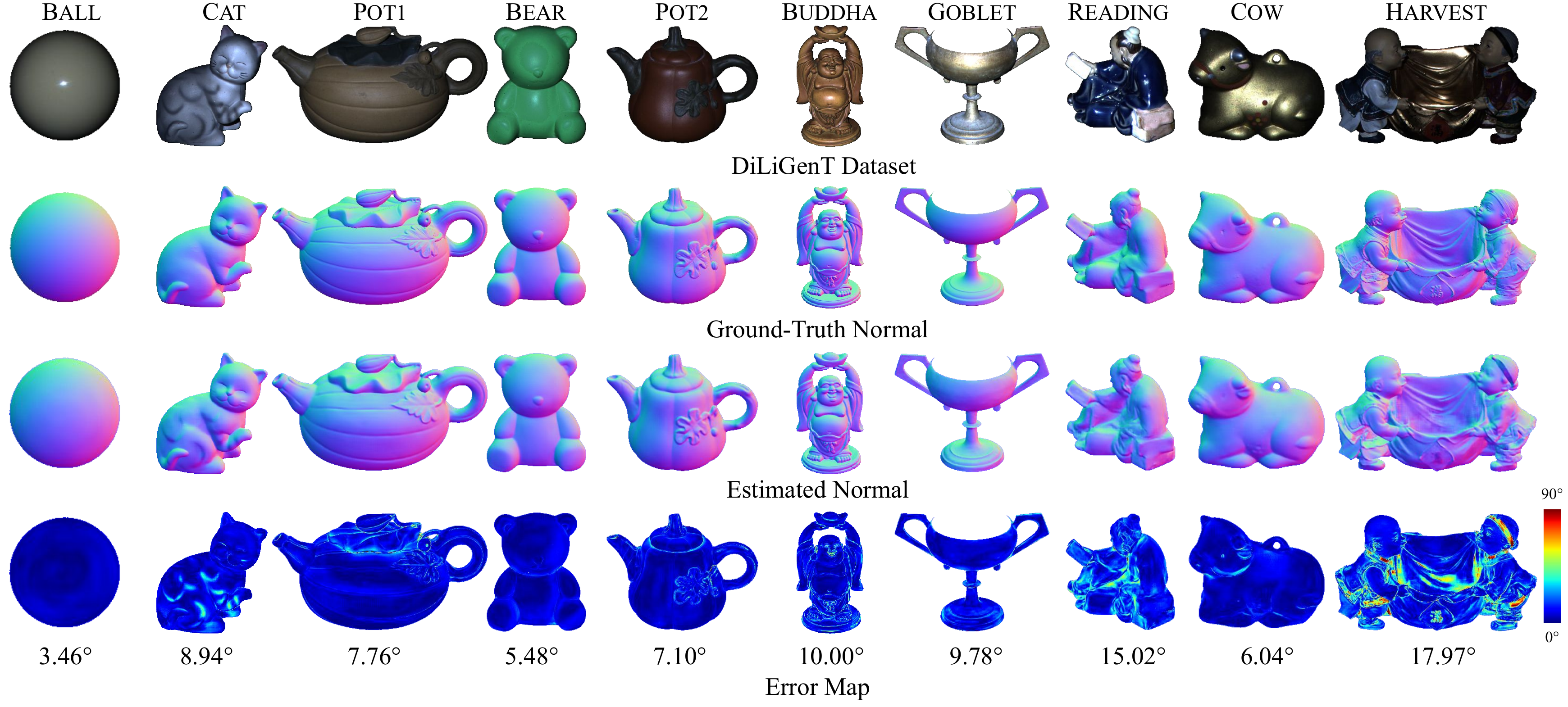}
}
\caption{\small  (a) Training curve of the normal estimation network.(b) Qualitative surface normal results on the DiLiGenT benchmark. The bottom row demonstrates the angular error maps and mean angular errors of our results. 
}
\label{fig:normal_estimation_results}
\end{figure*}

\begin{table*}
\scriptsize
\centering
\resizebox{\textwidth}{!}
{
\begin{tabular}{ r *{10}{|c} | c  }
\hline
\rowcolor[gray]{0.70}
\textbf{Methods}$\downarrow$ $|$ \textbf{Dataset}  $\rightarrow$ & \textbf{Ball} & \textbf{Cat} & \textbf{Pot1} & \textbf{Bear} & \textbf{Pot2}& \textbf{Buddha} &\textbf{Goblet}&\textbf{Reading} & \textbf{Cow} & \textbf{Harvest} & \textbf{Average} \\
\hline
 Alldrin et al. (2007)\cite{alldrin2008photometric} & 7.27& 31.45& 18.37& 16.81 &49.16 & 32.81 & 46.54 & 53.65 & 54.72 & 61.70 & 37.25 \\
\hline
 Shi et al. (2010)\cite{inproceedings} & 8.90 & 19.84 & 16.68 & 11.98 & 50.68 & 15.54 & 48.79 & 26.93 & 22.73 & 73.86 & 29.59 \\
\hline
 Wu \& Tan (2013)\cite{wu2013calibrating} & 4.39 & 36.55 & 9.39 & 6.42 & 14.52 & 13.19 & 20.57 & 58.96 & 19.75 & 55.51 & 23.93 \\
\hline
 Lu et al. (2013)\cite{lu2015uncalibrated} & 22.43 & 25.01 & 32.82 & 15.44 & 20.57 & 25.76 & 29.16 & 48.16 & 22.53 & 34.45 & 27.63 \\
\hline
 Papadh. et al. (2014)\cite{papadhimitri2014closed} & 4.77 & 9.54 & 9.51 & 9.07 & 15.90 & 14.92 & 29.93 & 24.18 & 19.53 & 29.21 & 16.66 \\
\hline
 Lu et al. (2017) \cite{lu2017symps} & 9.30 & 12.60 & 12.40 & 10.90 & 15.70 & 19.00 & 18.30 & 22.30 & 15.00 & 28.00 & 16.30 \\
\hline
\textbf{Ours} & \textbf{3.46}  & \textbf{8.94}  & \textbf{7.76} & \textbf{5.48} & \textbf{7.10} & \textbf{10.00} &  \textbf{9.78} & \textbf{15.02} & \textbf{6.04} & \textbf{17.97} & \textbf{9.15} \\
\hline
\end{tabular}
} 
\caption{\small Quantitative comparison with the traditional uncalibrated photometric stereo methods on DiLiGenT benchmark. Our searched architecture estimates accurate surface normals of the object with general reflectance property.}
\label{tab:uncalibrated_classical}
\end{table*}

\begin{figure*}[t]
\centering
\subfigure[\label{fig:cow_result_comparison_single} BEAR]{\includegraphics[ width=0.31\textwidth, height=0.12\textwidth]{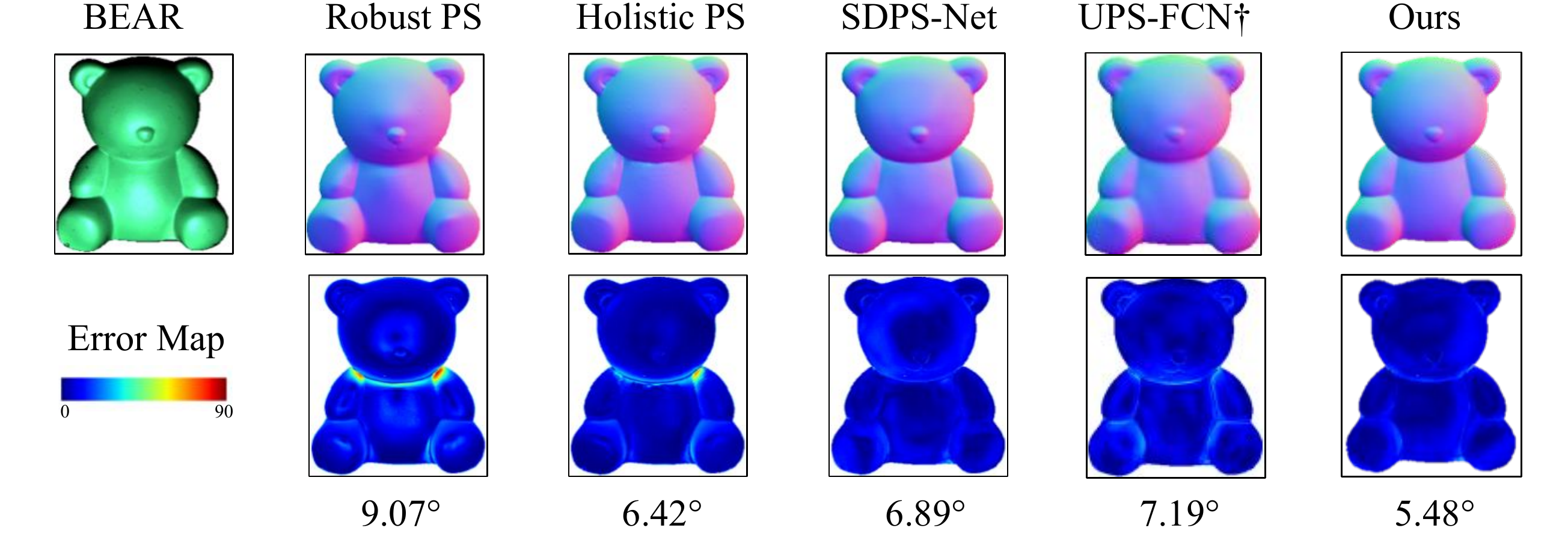}}%
~~\subfigure[\label{fig:goblet_result_comparison_single} GOBLET]{\includegraphics[ width=0.31\textwidth, height=0.12\textwidth]{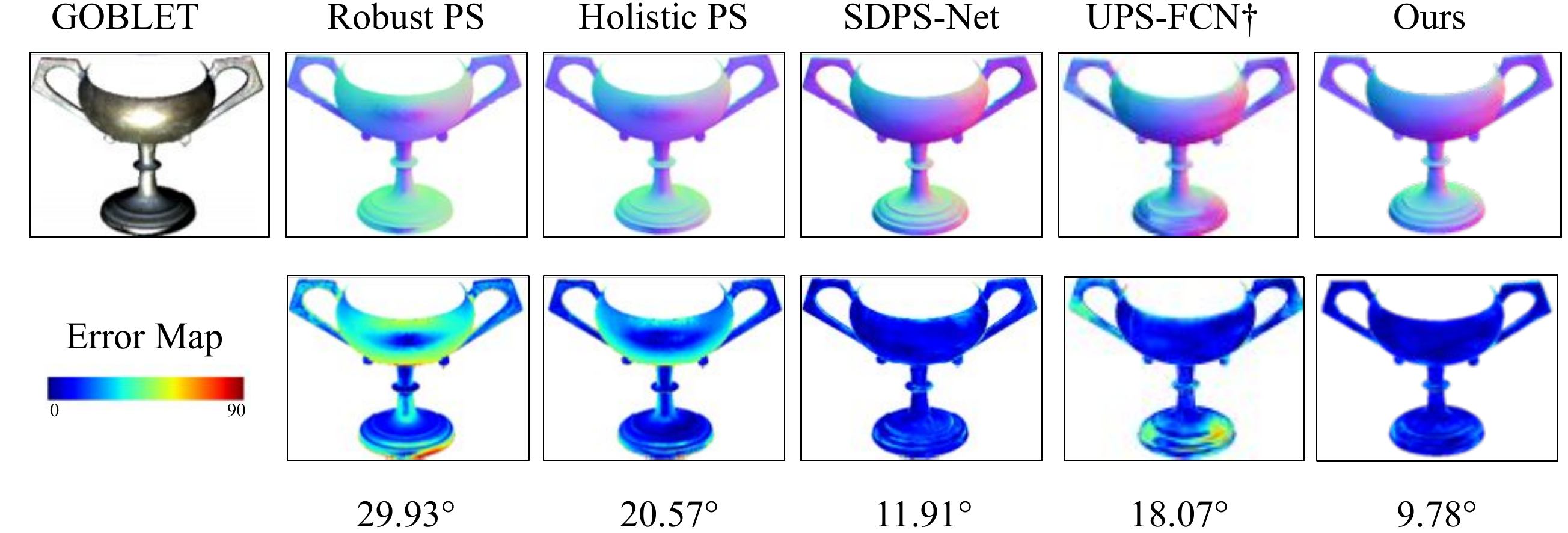}}%
~~\subfigure[\label{fig:pot2_result_comparison_single} POT2]{\includegraphics[ width=0.31\textwidth, height=0.12\textwidth]{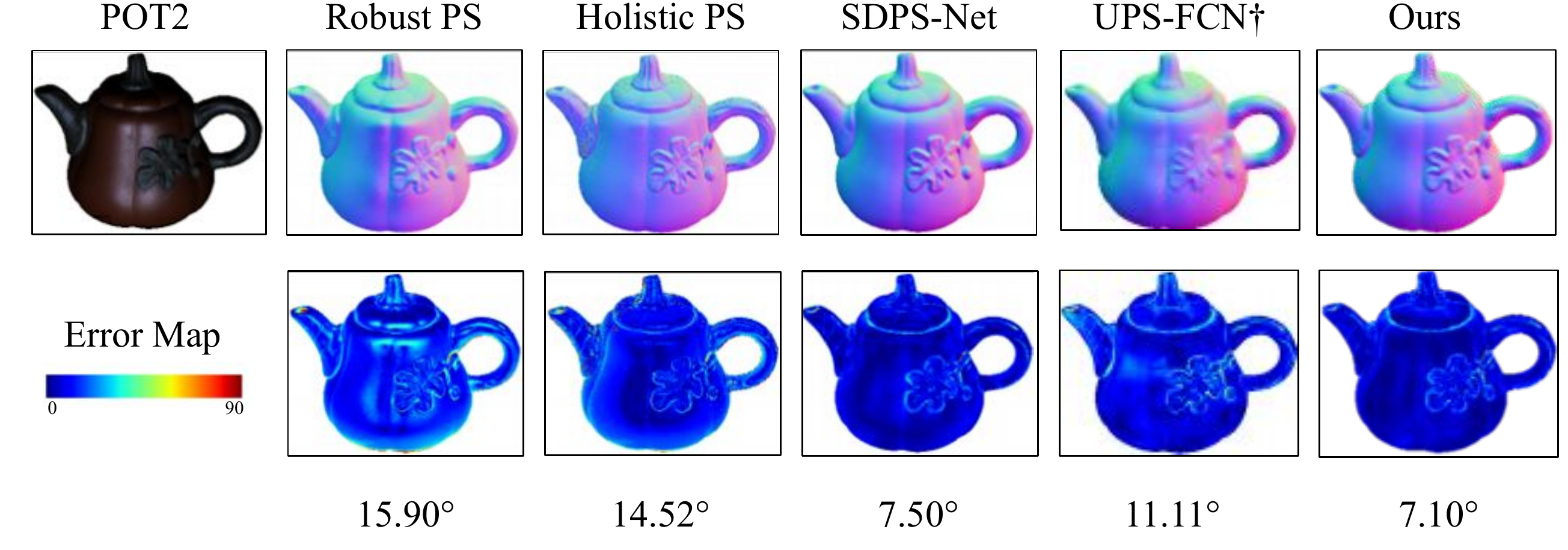}}
\caption{\small Visual comparison against Robust PS \cite{papadhimitri2014closed}, Holistic PS\cite{wu2013calibrating}, SDPS-Net \cite{chen2019self} and UPS-FCN \cite{chen2018psfcn} on (a) BEAR (b) GOBLET and (c) POT2 object from DiLiGenT dataset. The statistics show the superiority of our searched architecture.}
\label{fig:normal_estimation_results_single}
\end{figure*}

\subsubsection{Test Set Details.}\label{sss:test_set_details}
We tested our networks on the recently proposed DiLiGenT PS dataset \cite{shi2019tpami}. It consists of 10 real-world objects, with images captured by 96 LED light sources. It provides ground-truth normals and calibrated light directions making it an ideal dataset for evaluation. Following Chen \etal \cite{chen2018psfcn}, we use 96 images per object at $128 \times 128$ resolution to test our light calibration and normal estimation network.

\subsection{Implementation Details}
The proposed method is implemented with Python 3.6, and PyTorch 1.1 \cite{paszke2017automatic}. For both networks, we employ the same optimizer, learning rate, and weight decay settings. The architecture parameters $\alpha$ and the network weights $\omega$ are optimized using Adam \cite{kingma2017adam}.
During the architecture search phase, the optimizer is initialized with the learning rate $\eta_{alpha}= 3 \times 10^{-4}$, momentum $\beta = (0.5, 0,999)$ and weight decay of $1 \times 10^{-3}$.
At model train time, the optimizer is initialized with the learning rate $\eta_{w}= 5 \times 10^{-4}$, momentum $\beta = (0.5, 0,999)$ and weight decay of $3 \times 
10^{-4}$. We conducted all the experiments on a computer with a single NVIDIA GPU with 12GB of RAM.


 We search for two types of cells, namely normal cell and reduction cell. We use the loss function defined in Eq.\eqref{eq:lcnet_loss} and Eq.\eqref{eq:normal_loss} during search phase to recover optimal cells for each network independently. Fig.\ref{fig:LCNet_pipeline} and Fig.\ref{fig:NENet_pipeline} show the light calibration and the normal estimation backbone and its searchable parts, respectively. For light calibration network, we have two searchable blocks \textit{(i)} Feature block and \textit{(ii)} Classification block. Here, we design our feature block using three normal cells, two reduction cells, and the classification block using one normal cell and one reduction cell. Similarly, we have two searchable blocks \textit{(i)} Feature block and \textit{(ii)} Regressor block for normal estimation network. Here, the feature block comprises three normal cells and two reduction cells, while the regressor block is composed of three normal cells. To construct the network design for searchable blocks, each normal cell is concatenated sequentially to the reduction cell in order. We use 3 epochs to search architecture for each network.

\begin{table*}[t]
\scriptsize
\centering
\resizebox{\textwidth}{!}
{
\begin{tabular}{  r *{11}{|c} | c  }
\hline
\rowcolor[gray]{0.70}
\textbf{Methods} & \textbf{Params (M)} & \textbf{Ball} & \textbf{Cat} & \textbf{Pot1} & \textbf{Bear} & \textbf{Pot2}& \textbf{Buddha} &\textbf{Goblet}&\textbf{Reading} & \textbf{Cow} & \textbf{Harvest} & \textbf{Average} \\

 UPS-FCN$^\dagger$ (2018)\cite{chen2018psfcn} & 6.1 & 3.96 & 12.16 & 11.13 & 7.19 & 11.11 & 13.06 & 18.07 & 20.46 & 11.84 & 27.22 & 13.62 \\
\hline
 SDPS-Net (2019) \cite{chen2019self}& 6.6 & 2.77 & 8.06 & 8.14 & 6.89 & 7.50 & 8.97 & 11.91 & \textbf{14.90} & 8.48 & 17.43 & 9.51\\
\hline
GCNet (2020) \cite{chen2020learned} + PS-FCN \cite{chen2018psfcn} & 6.8 & \textbf{2.50} & \textbf{7.90} & \textbf{7.20} & 5.60 & \textbf{7.10} & \textbf{8.60} & 9.60 & \textbf{14.90} & 7.80 &  \textbf{16.20} & \textbf{ 8.70}\\
\hline
Kaya et al. (2021) \cite{kaya2020uncalibrated} & 8.1 & 3.78 & 7.91 & 8.75 & 5.96 & 10.17 & 13.14 & 11.94 & 18.22 & 10.85 & 25.49 & 11.62 \\
\hline

Ours (w/o auxiliary) &  \textbf{4.4} &  4.86 & 9.79 & 9.98 & \textbf{4.97} & 8.95 & 10.29 & \textbf{9.46} & 15.59 & 8.06 & 18.20 & 9.98 \\
\hline

\textbf{Ours} & \textbf{4.4} & 
3.46  & 8.94  &
\textcolor{blue}{\textbf{7.76}} & 
\textcolor{blue}{\textbf{5.48}} & \textbf{7.10} & \textcolor{blue}{\textbf{10.00}} &  9.78 & 15.02 & \textbf{6.04} & 
17.97 & \textcolor{blue}{\textbf{9.15}} \\

\hline

\end{tabular}
}
\caption{\small Quantitative comparison of deep uncalibrated photometric stereo methods on DiLiGenT benchmark \cite{shi2016benchmark}. Our searched architecture on average provides results that are better compared to other deep networks not only in surface orientation accuracy (MAE) but also in model size. The blue show the statistics where our method has the second best performance. We used deeper version of UPS-FCN \cite{chen2018psfcn}.}
\label{tab:uncalibrated_deep}
\end{table*}

At train time, we regularize the normal estimation network loss function using the concept of auxiliary tower \cite{liu2018darts} for performance gain. Consequently, we modify its loss function at train time as follows:
\begin{equation}
    \mathscr{L}_{normal} = \frac{1}{m} \sum_i^{m}(1-\boldsymbol{\tilde{n}}_i^T \boldsymbol{n}_i) + \lambda_{aux} \frac{1}{m} \sum_i^{m}(1-\boldsymbol{\hat{n}}_i^T \boldsymbol{n}_i)
    \label{eq:normal_loss_auxiliary}
\end{equation}
where, $\lambda_{aux}$ is a regularization parameter, and $\boldsymbol{\hat{n}}_i$ is the output surface normal at pixel $i$ due to auxiliary tower. We set $\lambda_{aux} = 0.4$. We observed that the auxiliary tower improves the performance of the normal estimation network. It can be argued that a similar regularizer could be used for the light calibration network. However, in that case, we have to incorporate that regularizer for each image independently, which can be computationally expensive. Fig.\ref{fig:LCNet_training_curve} and Fig.\ref{fig:NENet_training_curve} show the training curve for the light calibration and normal estimation network respectively. We trained the light calibration and normal estimation networks for six and three epochs, respectively for inference. 


\subsection{Qualitative and Quantitative Evaluation}

\formattedparagraph{Evaluation Metric.}
To measure the accuracy of the estimated light directions and surface normals, we adopt the standard mean angular error (MAE) metric as follows:

\begin{equation}
\small
\text{MAE}_{light} = \frac{180}{\pi} \frac{1}{n} \sum_{i}^{n} \mathrm{arccos}(\boldsymbol{\tilde{\ell}}_i^T \boldsymbol{\ell}_i) 
\end{equation} 
\begin{equation}
\small
\text{MAE}_{normal} = \frac{180}{\pi} \frac{1}{m} \sum_{i}^{m} \mathrm{arccos}(\boldsymbol{\tilde{n}}_i^T \boldsymbol{n}_i) 
\end{equation} 
where, $n$ is the number of images, and $m$ is the number of object pixels. $\boldsymbol{\tilde{\ell}}_i$ and $\boldsymbol{\ell}_i$ denote the estimated and ground-truth light directions. Similarly, $\boldsymbol{\tilde{n}}_i$ and $\boldsymbol{n}_i$ denote the estimated and ground-truth surface normals. As the auxiliary tower is not used at test time, we define metrics using $\boldsymbol{\tilde{n}}_i$. Following previous works \cite{chen2019self, chen2018psfcn}, we report MAE in degrees. 


Unlike light directions and surface normals, light intensity can only be estimated up to a scale factor. For this reason, instead of using the exact intensity values for evaluation, we use a scale-invariant relative error metric \cite{chen2019self}:

\begin{equation}
    \text{E}_{err} = \frac{1}{n} \sum_{i}^{n} \left( \frac{|s\tilde{e}_i - e_i|}{e_i} \right)
\end{equation}
Here, $\tilde{e}_i$ and $e_i$ are the estimated and ground-truth light intensities, respectively with $s$ as the scale factor. Following Chen \etal \cite{chen2020deep}, we solve ${\operatorname{argmin}}_s \sum_{i}^{n}(s\tilde{e}_i - e_i)^2 $ using the least squares to compute $s$ for intensity evaluation. 
\subsubsection{Inference}
Once optimal architectures are obtained, we train these networks for inference. We test their performance using the defined metric on the Test set. 
For each test object, we first feed the object images at $128 \times 128$ resolution to the light calibration network to predict the light directions and intensities. Then, we use the images and estimated light sources as input to the normal estimation network to predict the surface normals. Visual diagram of the optimal cell architectures is provided in the supplementary material.

\smallskip
\formattedparagraph{(a) Performance of Light Calibration Network.}
To show the validity of our searched light calibration network, we compared its performance on DiLiGenT ground-truth light direction and intensity. Fig.\ref{fig:light_maps} shows the quantitative and qualitative results obtained using our network. Concretely, it provides light directions $\text{MAE}_{light}$ and intensity error ($\text{E}_\textrm{err}$) for all object categories. The results indicate that the searched light calibration network can reliably predict light source direction and intensity from images of object with complex surface profile and different material properties.  

\smallskip
\formattedparagraph{(b) Comparison of Surface Normal Accuracy.} 
We documented the performance comparison of our approach against the traditional uncalibrated photometric stereo methods in Table \ref{tab:uncalibrated_classical}.
The statistics show that our method performs significantly better than such uncalibrated approaches for all the object categories. That is because we don't explicitly rely on BRDF model assumptions and the well-known matrix factorization approach. Instead, our work exploits the benefit of the deep neural network to handle complicated BRDF problems by learning from data.  Rather than using matrix factorization, our work independently learns to estimate light from data and use it to solve surface normals. 

Further, we compared our method with the state-of-the-art deep uncalibrated PS methods.
Table \ref{tab:uncalibrated_deep} shows that our method achieves competitive results with an average $\text{MAE}_{normal}$ of ${9.15}^{\circ}$, having the second best performance overall. The best performing method 
\cite{chen2020learned} uses a four-stage cascade structure, making it complex and deep. On the contrary, our searched architecture is light and it can achieve such accuracy with 2.4M fewer parameters. Fig.\ref{fig:normal_estimation_results_single} provides additional visual comparison of our results with several other approaches from the literature  \cite{papadhimitri2014closed, wu2013calibrating, chen2019self, chen2018psfcn}. Table \ref{tab:uncalibrated_deep} also shows the benefit of using an auxiliary tower at train time (see supplementary for more details and results).

\begin{figure}[h]
\centering
\subfigure[\label{fig:LCNET_numberofimages}Light Calibration Error]{\includegraphics[width=0.23\textwidth, height=0.22\textwidth]{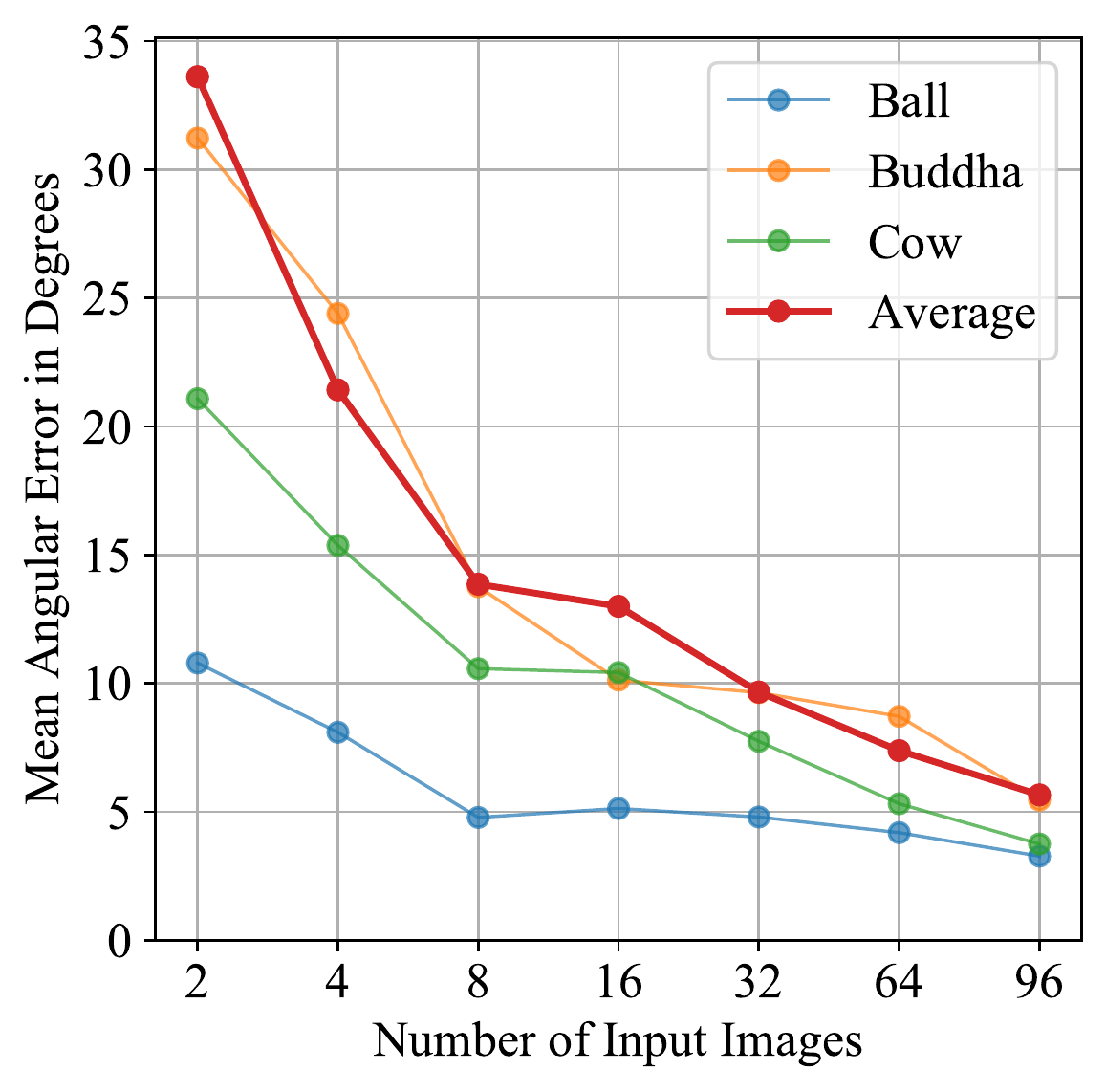}}~~
\subfigure[\label{fig:NENET_numberofimages}Surface Normal Error]{\includegraphics[ width=0.23\textwidth,height=0.22\textwidth ]{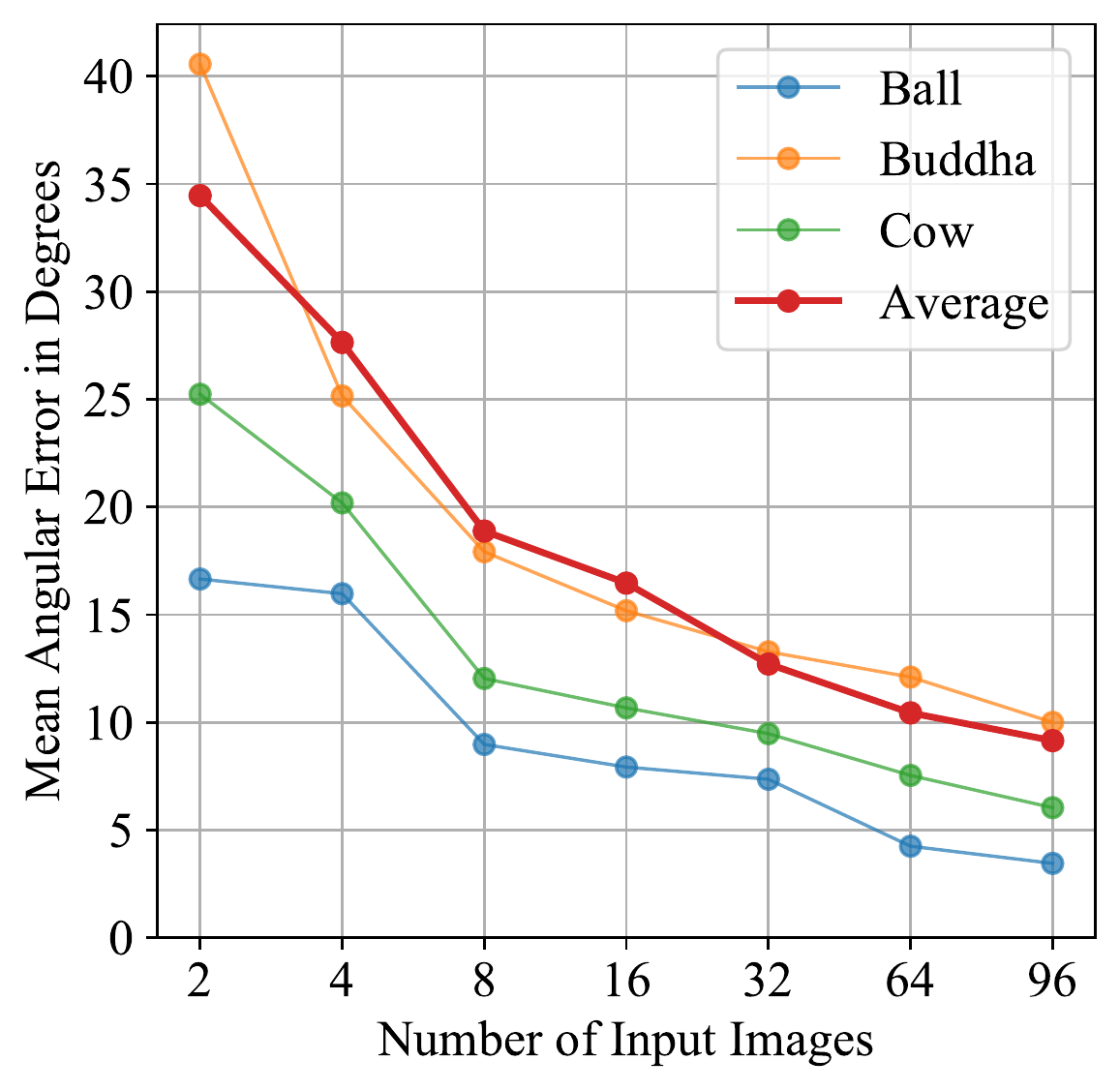}}~~
\caption{\small Variation in MAE w.r.t the change in the number of input images at test time. Observation with (a) light calibration and (b) normal estimation network, respectively.}
\label{fig:mae_vs_test_images}
\end{figure}

\formattedparagraph{(c) Ablation Study.}
\textit{(i) Analysing the performance with the change in number of input images at test time.}
Our light calibration and normal estimation network can work with an arbitrary number of input images at test time. In this experiment, we analyse how the number of images affects the accuracy of the estimated lighting and surface normals. Fig. \ref{fig:LCNET_numberofimages} and \ref{fig:NENET_numberofimages} show the variation in the mean angular error with different number of images. As expected, the error decreases as we increase the number of images. Of course, feeding more images allows the networks to extract more information, and therefore, the best results are obtained by using all 96 images provided by the DiLiGenT dataset \cite{shi2016benchmark}. For more experimental results, ablations and visualizations, refer to the supplementary material.

\section{Conclusion}

\noindent
In this paper, we demonstrated the effectiveness of applying differentiable NAS to deep uncalibrated PS. Though using the existing differentiable NAS framework directly to our problem is not straightforward, we showed that we could successfully utilize NAS  provided PS-specific constraints are well satisfied during the search, train, and test time. We search for an optimal light calibration network and normal estimation network using the one-shot NAS method by leveraging hand-crafted deep neural network design knowledge and fixing some of the layers or operations to account for the PS-specific constraints. The architecture we discover is lightweight, and it provides comparable or better accuracy than the existing deep uncalibrated PS methods. 

\smallskip
\formattedparagraph{Acknowledgement.} {{This work was funded by Focused Research Award from Google (CVL, ETH 2019-HE-318, 2019-HE-323).}}

{\small
\bibliographystyle{ieee_fullname}
\bibliography{arxiv_version}
}


\end{document}